%% file: main.tex
\documentclass[lettersize,journal]{IEEEtran}
\usepackage{amsmath,amsfonts}
\usepackage{bbm}
\usepackage{algorithmic}
\usepackage{algorithm}
\usepackage{array}
\usepackage[caption=false,font=normalsize,labelfont=sf,textfont=sf]{subfig}
\usepackage{textcomp}
\usepackage{stfloats}
\usepackage{url}
\usepackage{verbatim}
\usepackage{graphicx}
\usepackage{cite}
\usepackage{multirow}
\usepackage{booktabs} 
\usepackage[colorlinks=true]{hyperref}
\def\netName{DeMo}
\def\netNameNew{DeMo++}

\def\netNameetwoeNew{DeMo-E2E++}

\newcommand{\ext}[1]{\textcolor{black}{#1}}

\begin{document}


\title{\netNameNew: Motion Decoupling for Autonomous Driving}

\author{
Bozhou Zhang*, Nan Song*, Xiatian Zhu, Li Zhang
\thanks{
* Equal contribution.

Li Zhang is the corresponding author (lizhangfd@fudan.edu.cn). Bozhou Zhang, Nan Song and Li Zhang are with the School of Data Science, Fudan University. Xiatian Zhu is with the University of Surrey.
}
}


\markboth{Journal of \LaTeX\ Class Files,~Vol.~14, No.~8, August~2021}%
{Shell \MakeLowercase{\textit{et al.}}: A Sample Article Using IEEEtran.cls for IEEE Journals}


\maketitle


\input{section/0_abstract}
\input{section/1_introduction}

\input{section/2_relatedwork}

\input{section/3_method}
\input{section/4_experiment}

\input{section/5_conclusion}




\bibliographystyle{IEEEtran}
\bibliography{main}



\begin{IEEEbiography}
[{\includegraphics[width=0in,height=0in,clip,keepaspectratio]{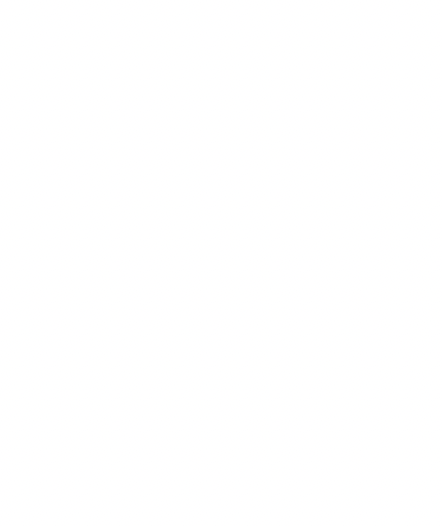}}]{Bozhou Zhang}
received his BEng degree from Beihang University. He is now a PhD student in the School of Data Science, Fudan University under the supervision of Prof. Li Zhang. 
His research interests include embodied intelligence and autonomous driving.
\end{IEEEbiography}
\vspace{-10mm}

\begin{IEEEbiography}
[{\includegraphics[width=0in,height=0in,clip,keepaspectratio]{figure/authors/sample.png}}]{Nan Song}
is currently a PhD student at Fudan University. His research interests
include autonomous driving and machine learning.
\end{IEEEbiography}
\vspace{-10mm}

\begin{IEEEbiography}
[{\includegraphics[width=0in,height=0in,clip,keepaspectratio]{figure/authors/sample.png}}]{Xiatian Zhu}
received the PhD degree from Queen
Mary University of London. He is a senior lecturer
 at the University of Surrey. 
He won the Sullivan Doctoral Thesis Prize
2016. 
His research interests
include computer vision and machine learning.
\end{IEEEbiography}
\vspace{-10mm}

\begin{IEEEbiography}
[{\includegraphics[width=0in,height=0in,clip,keepaspectratio]{figure/authors/sample.png}}]{Li Zhang}
received the PhD degree in computer
science from the Queen Mary University of London.
He is now a full professor at Fudan University.
Previously, he was a postdoctoral research fellow
at the University of Oxford. His research interests
include computer vision and embodied AI. 
\end{IEEEbiography}
\vspace{-10mm}


\end{document}

%% file: section/0_abstract.tex
\begin{abstract}
Motion forecasting and planning are tasked with estimating the trajectories of traffic agents and the ego vehicle, respectively, to ensure the safety and efficiency of autonomous driving systems in dynamically changing environments. State-of-the-art methods typically adopt a \textit{one-query-one-trajectory} paradigm, where each query corresponds to a unique trajectory for predicting multi-mode trajectories.
While this paradigm can produce diverse motion intentions, it often falls short in modeling the intricate spatiotemporal evolution of trajectories, which can lead to collisions or suboptimal outcomes.
To overcome this limitation, we propose \textit{\netNameNew{}}, a framework that decouples motion estimation into two distinct components: {\em holistic motion intentions} to capture the diverse potential directions of movement,
and {\em fine spatiotemporal states} to track the agent's dynamic progress within the scene and enable a self-refinement capability.
Further, we introduce a cross-scene trajectory interaction mechanism to explore the relationships between motions in adjacent scenes.
This allows \netNameNew{} to comprehensively model both the diversity of motion intentions and the spatiotemporal evolution of each trajectory.
To effectively implement this framework, we developed a hybrid model combining Attention and Mamba. This architecture leverages the strengths of both mechanisms for efficient scene information aggregation and precise trajectory state sequence modeling. Extensive experiments demonstrate that \netNameNew{} achieves state-of-the-art performance across various benchmarks, including motion forecasting (Argoverse 2 and nuScenes), motion planning (nuPlan), and end-to-end planning (NAVSIM).
Our code is available at \url{https://github.com/fudan-zvg/DeMo}.

\end{abstract}


%% file: section/1_introduction.tex
\section{Introduction}
\label{introduction}

Motion forecasting~\cite{survey2022, waymo, argoverse2} empowers self-driving vehicles to anticipate how surrounding agents will move and influence the ego vehicle, based on which motion planning~\cite{carla, navsim, nuplan} needs to generate feasible driving trajectories for the ego vehicle.
These tasks are critical for maintaining safety and dependability, enabling vehicles to comprehend the dynamics of driving environments and make calculated decisions. The challenges and complexities of these tasks arise from various factors, including unpredictable road conditions, varied movement patterns of traffic participants, and the necessity to simultaneously analyze the states of observed agents along with the road maps.

\input{figure/fig_first}

The research community has witnessed significant progress in the representation of driving scenes~\cite{vectornet,LaneGCN,SceneTransformer,plantf} and the paradigm of trajectory decoding~\cite{densetnt,eda,mtr,qcnet, diffusiondrive, goalflow,diffusionplanner,betop}. These methods have achieved substantial advancements in estimation accuracy, primarily following a certain pattern inspired by detection~\cite{detr,DAB-DETR}, \emph{i.e.}, the one-query-one-trajectory paradigm~\cite{mtr,qcnet,diffusiondrive, goalflow}. This paradigm utilizes several queries to represent different estimated trajectories, as shown in Figure~\ref{fig:first} (a), enabling distinct motion intentions. Although effective, these approaches can only approximately provide a direction and collect surroundings to generate various trajectory waypoints in a one-shot fashion, overlooking the detailed relationships with scenes. The lack of concrete representation for trajectories and comprehensive spatiotemporal interactions with the surrounding environment and among each other might lead to a decline in accuracy and consistency across varying time steps.

To solve this problem, we propose a novel framework dubbed {\bf\netNameNew{}}, which provides a structured representation of multi-mode\footnote{Here we use the term ``multi-modal" to describe the input data, and ``multi-mode" to refer to diverse motion forecasting and motion planning decisions.} trajectories. Specifically, we decouple motion estimation into two facets: besides the original motion modes to capture different directional intentions (Figure~\ref{fig:first} (a)), we introduce the spatiotemporal states for future trajectories to track the agent's dynamic motion progress across various space positions and time steps (Figure~\ref{fig:first} (b)). This approach allows us to achieve a comprehensive motion representation within our framework (Figure~\ref{fig:first} (c)). Mode intentions and states are processed using the Mode Localization Module and the State Consistency Module, respectively. Subsequently, these two types of representations are integrated by our Hybrid Coupling Module to achieve a comprehensive modeling of future trajectories. Due to the sequential nature of trajectory states, Mamba~\cite{mamba} is particularly selected for modeling the temporal consistency of dynamic states. Therefore, we utilize a combination of Attention and Mamba in our modules to effectively and efficiently aggregate global information and model state sequences, leveraging the strengths of both techniques.

\ext{
With this decoupled trajectory representation, we further exploit the potential of accurate and continuous motion modeling in real-world driving scenarios. Considering that this paradigm models trajectories based on global intentions and local states, we enhance these two motion representations by enabling cross-scene intention interactions and by refining trajectory predictions using state anchors. The former maintains trajectory consistency according to intention similarity across scenes, reinforcing continuous driving in real-world scenarios; While the latter is utilized to refine the current predictions through state anchor-based scene interaction, which can enhance accuracy and mitigate unreasonable predictions, such as collisions. 
}

Our \textbf{contributions} are summarized as follows: 
{\bf (i)} We propose a motion forecasting and motion planning framework, \netNameNew{}, which decouples multi-mode trajectory representations into motion modes and dynamic states to separately capture directional intentions and movement progress. 
\ext{
{\bf (ii)} We further incorporate cross-scene intention interaction and state anchor-based refinement, fully unlocking the potential of the decoupling paradigm.
}
\ext{
{\bf (iii)} Extensive experiments on the motion forecasting benchmarks Argoverse 2 and nuScenes, the motion planning benchmark nuPlan, and the end-to-end planning benchmark NAVSIM demonstrate that \netNameNew{} achieves state-of-the-art performance.
}

\ext{
Our preliminary works, \textbf{\netName{}~\cite{demo}} and \textbf{RealMotion~\cite{RealMotion}}, have both been presented at NeurIPS 2024.
This journal submission further enhances the motion decoupling paradigm through novel module and architectural designs.
(i) We advance the motion decoupling strategy by introducing cross-scene intention interaction and state anchor-based refinement.
(ii) We extend our application to the motion planning task, focusing on the predicted trajectories for the ego vehicle. 
(iii) We incorporate raw sensor data and adapt our model to end-to-end autonomous driving, covering diverse driving tasks from perception and prediction to planning. (iv) We conduct more extensive ablation studies, providing a comprehensive analysis of the performance improvement and exploring the scalability of our framework.
}

%% file: figure/fig_first.tex

\begin{figure}[t!]
\centering
\includegraphics[width=0.48\textwidth]{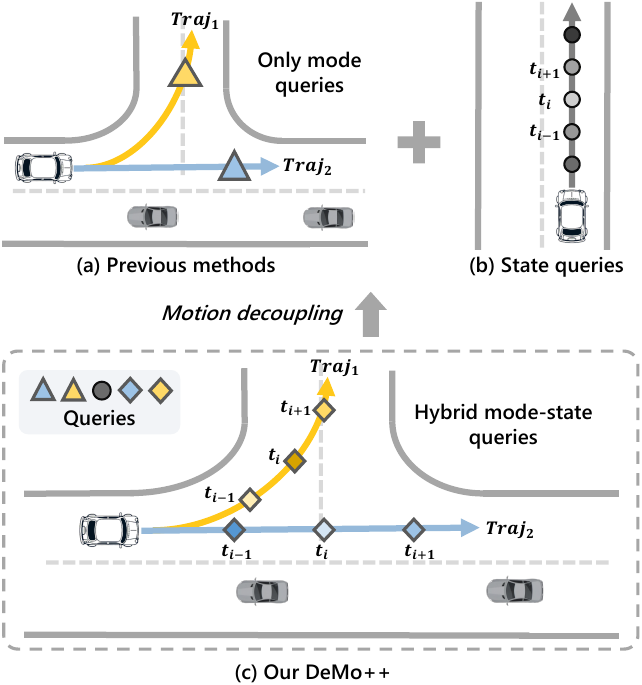}
\caption{
Conceptual illustration of future trajectory representation.
{\bf(a)} Previous methods use only one mode query for each trajectory.
{\bf(c)} Our approach adopts a novel decoupled query strategy, which introduces {\bf(b)} state queries in addition to mode queries to represent multi-mode trajectories.
}
\label{fig:first}
\end{figure}

%% file: section/2_relatedwork.tex
\section{Related work}

\paragraph{Motion forecasting}

In recent advancements in autonomous driving, it is critical to effectively predict the movements of relevant agents by accurately representing scene components. Traditional methods~\cite{multipath,home,covernet} transformed driving scenarios into image formats and used conventional convolutional networks for scene context encoding. However, these techniques often failed to sufficiently capture intricate structural details. This challenge has led to the adoption of vectorized scene representations~\cite{densetnt,multipath++,tnt,hivt}, exemplified by the introduction of VectorNet~\cite{vectornet}. Additionally, graph-based structures are also widely utilized to represent the relationships between agents and their environments~\cite{pgp,gohome,noname,hdgt,LaneGCN,fjmp,lanercnn}.

Existing methodologies have delved into a variety of frameworks to predict multi-mode future trajectories given the scene features. Initially, prediction techniques were centered on goal-oriented methods~\cite{densetnt,DSP} or employed probability heatmaps to sample trajectories~\cite{home,gohome}. However, contemporary strategies, such as MTR~\cite{mtr} and QCNet~\cite{qcnet}, among others~\cite{mmTransformer,wayformer,SceneTransformer,HPTR}, utilize Transformer~\cite{attention} models to analyze relationships within the scene. Additionally, the introduction of novel paradigms such as pre-training~\cite{Traj-mae,Forecast-mae,sept}, historical prediction design~\cite{t4p,hpnet}, GPT-style next-token prediction~\cite{trajeglish,motionlm}, and post-refinement~\cite{R-Pred,smartrefine} in some techniques has led to remarkable advancements in performance.

Furthermore, the advancements in multi-agent forecasting aim to enhance the applicability of predicted trajectories for various agents in real-world scenarios. Several approaches~\cite{thomas,mtr++,hivt} follow an agent-centric model, where trajectories are forecasted individually for each agent, a process that might be slow. On the other hand, alternative approaches~\cite{SceneTransformer,qcnext} utilize a scene-centric model that allows for simultaneous forecasting across all agents, introducing an innovative approach to trajectory prediction.

Inspired by the progress in object detection and motivated by its significant success~\cite{detr,DAB-DETR}, mainstream methods~\cite{eda,mtr,hpnet,qcnet} in motion forecasting have adopted a one-query-one-trajectory paradigm to achieve high performance in motion forecasting benchmarks~\cite{argoverse,waymo,argoverse2,nuscenes}. These methods leverage transformers to model the relationship between each trajectory query and its environment, but they lack detailed trajectory representations. To address this limitation, we propose decoupled mode queries and state queries to enable a more detailed and comprehensive representation of multi-mode trajectories.

\paragraph{\ext{Motion planning}}

\ext{
After understanding the driving environment and obtaining the upstream perception and forecasting results, motion planning is tasked with generating feasible driving trajectories for the ego vehicle. One mainstream research direction~\cite{nuplan} focuses exclusively on planning, eliminating the perception requirements and simplifying driving scenes by representing them with agent trajectories and an HD map. In this task setting, rule-based models~\cite{idm, PDM}, which rely on strict traffic rule constraints, still play a crucial role. Nevertheless, learning-based methods have emerged and have surpassed traditional approaches in recent years. For instance, PlanTF~\cite{plantf} and PLUTO~\cite{pluto} improve the model architecture and training strategies for planning, effectively alleviating the limitations of imitation-based methods. In addition, BeTopNet~\cite{betop} explores the topological relationships among scene elements and explicitly represents the behavioral topology, which further enhances planning performance.
}

\paragraph{\ext{End-to-end autonomous driving}}
\ext{
The integrated end-to-end autonomous driving frameworks~\cite{UniAD,vad,sparsedrive,diffusiondrive,bridgead} have also attracted increasing attention. These frameworks take raw sensor data as input and encompass various driving tasks, ranging from perception~\cite{bevformer,streampetr}, and prediction~\cite{mtr,qcnet}, to planning~\cite{plantf,diffusionplanner}.
Early methods~\cite{TransFuser,stp3,driveadapter,thinktwice} tended to bypass intermediate tasks and directly perform planning based on sensor data for both open-loop~\cite{nuscenes} and closed-loop~\cite{carla} tasks.
UniAD~\cite{UniAD} pioneered the integration of perception, prediction, and planning into a unified framework with a straightforward Transformer architecture. It adopts a planning-oriented approach to optimize the overall pipeline, achieving remarkable performance across all tasks.
Following this design principle, VAD~\cite{vad} introduces a vectorized representation and simplifies the task structure, improving the efficiency of end-to-end systems. The follow-up work~\cite{vadv2} further presents a probabilistic planning paradigm equipped with a large vocabulary. In addition, sparse frameworks~\cite{sparsedrive} have also been explored to better utilize temporal information and enhance inference efficiency.
Several other studies~\cite{LAW,SSR} simplify the complex end-to-end pipelines by employing self-supervised learning.
}

\ext{
Recently, research has increasingly focused on more challenging end-to-end planning benchmarks~\cite{Bench2Drive,navsim}. In particular, DiffusionDrive~\cite{diffusiondrive} employs a diffusion policy with a truncation strategy to enable efficient and diverse planning. In contrast, GoalFlow~\cite{goalflow} focuses on achieving more precise planning performance by introducing flow matching and goal-point guidance into end-to-end frameworks. Inspired by the success of Large Language Models, DriveTransformer~\cite{drivetransformer} introduces a holistic Transformer architecture that aggregates all driving features for planning.
}

\paragraph{State space models}
Originally developed for modeling dynamic systems with state variables in fields such as control theory, state space models (SSMs) have emerged as promising alternatives to Transformers~\cite{attention} in sequence modeling, particularly due to their effectiveness in addressing attention complexity and capturing long-term dependencies. As SSMs have evolved~\cite{h3,s4,s5}, a new class termed Mamba~\cite{mamba}, which incorporates selection mechanisms and hardware-aware architectures, has recently demonstrated significant promise in long-sequence modeling. Several studies have explored Mamba's substantial potential across a range of fields, including natural language processing~\cite{densemamba,jamba} and computer vision~\cite{zigma,videomamba,motionmamba,vim}. Notably, in the vision domain, Mamba has demonstrated superior GPU efficiency and effectiveness compared to Transformers in tasks such as visual representation learning~\cite{vim}, video understanding~\cite{videomamba}, and human motion generation~\cite{motionmamba}. 
Building on these achievements, to the best of our knowledge, this is the first method to combine the strengths of Mamba with the mainstream Transformer-based architecture, achieving impressive performance in motion forecasting and planning.

%% file: section/3_method.tex
\section{Motion decoupling for motion forecasting and motion planning}

\label{methodology}
\input{figure/fig_main}

We present \textbf{\netNameNew{}}, which utilizes decoupled mode queries and state queries for directional intentions and dynamic states to predict future trajectories, as illustrated in Figure~\ref{fig:main}. We derive a hybrid architecture combining Attention and Mamba, along with two auxiliary losses for feature modeling. 
\ext{
To meet the high demands of precision and continuity in real-world scenarios, we further exploits the potential of our motion decoupling strategy. Building upon decoupled mode and state queries, we introduce cross-scene intention interaction to enhance motion continuity and state anchor-based refinement to improve estimation precision (Figure~\ref{fig:plus}).
}

\subsection{Problem formulation}
\label{sec:prob_form}

Given HD map and agents in the driving scenario, motion forecasting aims to predict the future trajectories for the interested agents. The HD map comprises several polylines of lanes or crossings, while agents are traffic participants like vehicles and pedestrians. To transform these elements into easily processable and learnable inputs, we utilize a popular vectorized representation following~\cite{vectornet,Forecast-mae,qcnet,mtr}. Specifically, the map ${\rm M} \in \mathbb{R}^{N_{\rm m} \times L \times C_{\rm m}}$ is generated by dividing each line into several shorter segments, where $N_{\rm m}$, $L$, and $C_{\rm m}$ denote the number of map polylines, divided segments, and feature channels, respectively. We represent the historical information of agents as ${\rm A} \in \mathbb{R}^{N_{\rm a} \times T_{\rm h} \times C_{\rm a}}$, where $N_{\rm a}$, $T_{\rm h}$, and $C_{\rm a}$ are the number of agents, historical timestamps, and motion states (e.g., position, heading angle, velocity). Additionally, the future trajectories ${A_{\rm m}} \in \mathbb{R}^{N_{\rm aoi} \times T_{\rm m} \times 2}$ for agents of interest are estimation objectives, with $N_{\rm aoi}$, $T_{\rm m}$ indicating the number of selected agents and the future timestamps, respectively.

\subsection{Scene context encoding}
Given the vectorized representations ${\rm A}$ for agents and ${\rm M}$ for HD map, we first employ individual encoders to process them separately. Specifically, we use a PointNet-based polyline encoder, as described in~\cite{Forecast-mae,mtr,mtr++}, to process the map representation ${\rm M}$, generating the map features $F_{\rm m} \in \mathbb{R}^{N_{\rm m} \times C}$. For the agents ${\rm A}$, we replace Transformer~\cite{attention} or RNN with several Unidirectional Mamba~\cite{mamba} blocks, which are more efficient and effective for sequence encoding, to aggregate the historical trajectory features $F_{\rm a} \in \mathbb{R}^{N_{\rm a} \times C}$ up to the current time. Subsequently, the scene context features $F_{\rm s} \in \mathbb{R}^{(N_{\rm a} + N_{\rm m}) \times C}$ are formed by concatenating them and further propagated to a Transformer encoder for intra-interaction learning. The overall process can be formulated as:
\begin{equation}
\begin{split}
F_{\rm m} &= {\rm PointNet}({\rm M}),\\ 
F_{\rm a} &= {\rm UniMamba}({\rm A}), \\
\quad F_{\rm s} &= {\rm Transformer}({\rm Concat}(F_{\rm a},\, F_{\rm m})).
\end{split}
\end{equation}

\subsection{Trajectory decoding with decoupled queries}
After obtaining the scene context features, we aim to decode multi-mode future trajectories for each interested agent based on our proposed decoupled queries. As illustrated in Figure~\ref{fig:main}, the decoder network comprises a State Consistency Module that enhances the consistency and accuracy of dynamic future state queries, a Mode Localization Module for learning distinct motion modes, and a Hybrid Coupling Module to integrate the decoupled queries and generate the final output. The detailed description of these components is provided in the following.

\paragraph{Dynamic state consistency}
Considering the recurrence and causality of the future trajectories ${A_{\rm m}}$, we propose to represent them as a series of dynamic states across various time steps, distinct yet interconnected. To preserve precise time information, the state queries ${Q_{\rm s}} \in \mathbb{R}^{N_{\rm aoi} \times T_{\rm s} \times C}$ are initialized with an MLP module for real-time differences. It is notable that the steps $T_{\rm s}$ can differ from $T_{\rm m}$ to balance the effectiveness and efficiency, especially when predicting long-term future trajectories or a higher frequency of future trajectories. The State Consistency Module is then employed to enhance the consistency of the state queries and aggregate the specific scene context, which can be formulated as follows:
\begin{equation}
\begin{split}
Q_{\rm s} &= {\rm MLP}([t_{1}, t_{2}, \cdots, t_{T_{\rm s}}]), \\
Q_{\rm s} &= {\rm MHA}({\rm Q} = Q_{\rm s}, {\rm K} = F_{\rm s}, {\rm V} = F_{\rm s}), \\
Q_{\rm s} &= {\rm BiMamba}(Q_{\rm s}).
\end{split}
\end{equation}

Specifically, cross-attention is first applied to enable state queries to interact with the scene context, followed by a Mamba block to model sequence relationships with linear-time complexity. Simultaneously, to account for the influences of rear state queries on the front ones, we adopt the bidirectional Mamba~\cite{videomamba,vim} for both forward and backward scanning. Additionally, a simple MLP module is utilized to decode the state queries $Q_{\rm s}$ into a single future trajectory for explicit supervision of time consistency.

\paragraph{Directional intention localization}
Mode queries ${Q_{\rm m}} \in \mathbb{R}^{N_{\rm aoi} \times K \times C}$ represent different motion modes, with each query responsible for decoding one of the $K$ trajectories. We utilize the Mode Localization Module to localize the potential directional intentions, as shown below:
\begin{equation}
\begin{split}
Q_{\rm m} &= {\rm MHA}({\rm Q} = Q_{\rm m}, {\rm K} = F_{\rm s}, {\rm V} = F_{\rm s}), \\
Q_{\rm m} &= {\rm MHA}({\rm Q} = Q_{\rm m}, {\rm K} = Q_{\rm m}, {\rm V} = Q_{\rm m}).
\end{split}
\end{equation}

For spatial motion learning, two Multi-Head Attention blocks are employed to enable interactions among mode queries and with the scene context. Additionally, we also employ simple MLPs to decode the future trajectories and probabilities. Similarly, we introduce another auxiliary supervision to endow mode queries with distinct motion intentions.

\paragraph{Hybrid query coupling}
To incorporate dynamic states and directional intentions, we simply add $Q_{\rm m}$ and $Q_{\rm s}$ together to form the hybrid spatiotemporal queries $Q_{\rm h} \in \mathbb{R}^{N_{\rm aoi} \times K \times T_{\rm s} \times C}$. Then, the Hybrid Coupling Module is utilized to further process $Q_{\rm h}$ and yield a comprehensive representation for future trajectories, as formulated below:
\begin{equation}
\begin{split}
Q_{\rm h} &= {\rm MHA}({\rm Q} = Q_{\rm h}, {\rm K} = F_{\rm s}, {\rm V} = F_{\rm s}), \\
Q_{\rm h} &= {\rm HybridMHA}({\rm Q} = Q_{\rm h}, {\rm K} = Q_{\rm h}, {\rm V} = Q_{\rm h}), \\
Q_{\rm h} &= {\rm ModeMHA}({\rm Q} = Q_{\rm h}, {\rm K} = Q_{\rm h}, {\rm V} = Q_{\rm h}), \\
Q_{\rm h} &= {\rm BiMamba}(Q_{\rm h}).
\end{split}
\end{equation}

Besides the Attention and Mamba modules for interaction with the scene context, among modes, and across time states, we additionally introduce a hybrid self-attention layer, which connects queries across both time and modes, boosting the diversity of predicted trajectories. The change in feature dimensions in this module is shown in Figure~\ref{fig:main} (c). The final predictions are generated by decoding the output $Q_{\rm h}$ into trajectory positions and probabilities with MLPs.

\input{figure/fig_plus}

\subsection{\ext{Cross-scene intention interaction}}

\ext{
In real-world scenarios, motion forecasting and planning are performed continuously as the ego vehicle moves forward, requiring motion intentions to maintain temporal coherence over time. Motivated by this, we introduce interactions between motion intentions across scenes to enhance temporal consistency and improve real-world applicability. Specifically, we reorganize snapshot-based datasets, such as Argoverse 2~\cite{argoverse2} and nuPlan~\cite{nuplan}, by converting them into sequential data. This is achieved by dividing each scene into sub-scenes, making these datasets more realistic in reflecting continuous driving behavior. We then \textit{apply our framework to each sub-scene and further introduce cross-scene intention interaction for both mode queries and state queries}.
}

\paragraph{\ext{Data reorganizing}}

\ext{
Snapshot-based datasets~\cite{argoverse, argoverse2, nuplan} consist of truncated scene samples that are independent of and irrelevant to each other, which conflicts with the nature of realistic driving scenarios. To address this issue, we reorganize the trajectories within each scene, transforming them into sequences using a sliding window technique with a fixed time step to better simulate continuous driving in the real world, as shown in Figure~\ref{fig:data}. The sliding window starts from the current step and moves backward in time, dividing the entire scene into sequential sub-scenes spanning from the past to the present. Each sub-scene contains both historical and future trajectory segments, analogous to the structure of the original scene. In this setting, the future segment retains the same length as in the original data, while the historical segment is slightly shorter. Additionally, for each sub-scene, we extract a local HD map within a specified range. Notably, we apply the same data processing pipeline to sub-scenes as described in Section~\ref{sec:prob_form}. This approach improves data utilization and supports more effective exploration of temporal information.
}

\paragraph{\ext{Mode query interaction}}

\ext{
In the continuous driving situation, motion intention should keep consecutive and consistent across scenes. Hence, we anticipate that historical mode features can affect and improve current motion intention. To achieve this, we adopt direct mode query interaction module according to trajectory similarity. The overall process is illustrated in Figure~\ref{fig:plus} (a). Specifically, we first decode the current and historical trajectories $Y_{\rm m}$ and $Y'_{\rm m}$ from the corresponding mode queries $Q_{\rm m}$ and $Q'_{\rm m}$. Considering that the trajectories are calculated based on respective local system, we then project the historical trajectories onto the current system, which can be formulated as:
\begin{equation}
Y'_{\rm m} = \mathcal{R}\cdot(Y'_{\rm m} - {y'}_{\rm m}^{\rm ori})^{\rm T}, 
\end{equation}
where $\mathcal{R}$ denotes the rotation matrix from historical system to current system. Besides, as the current position of agent frequently lies outside the historical predictions, the transformation with real position offsets might cause suboptimal similarity comparison. To alleviate this, we project historical trajectories based on the waypoints in the historical trajectories corresponding to the current time step, which is ${y'}_{\rm m}^{\rm ori}$. by which all projected trajectories pass through the current origin.
}

\ext{
After the projection, the historical and current trajectories that share overlapping segments are expected to have stronger correlations of mode features. To explicitly introduce this principle, we establish the interaction between current and historical mode queries through a lightweight Transformer module with Trajectory Embedding (TE) replacing the original Positional Embedding and modeling the geometric representations of trajectories. This procedure can be defined as follows:
\begin{equation}
Q_{\rm m} = {\rm Transformer}(Q_{\rm m} + {\rm TE}(Y_{\rm m}), \; Q'_{\rm m} + {\rm TE}(Y'_{\rm m})),
\end{equation}
where the Trajectory Embedding is computed through a MLP module to embed the flattened trajectories. Then, the updated mode queries $Q_{\rm m}$ are integrated into the hybrid queries, providing more accurate and temporally consistent motion intentions.
}

\paragraph{\ext{State query interaction}}
\ext{Benefiting from the dynamic state representation provided by state queries in our model, we maintain temporal consistency across scenes by enabling interactions between current and historical state queries. Following a process similar to that shown in Figure~\ref{fig:plus} (a), we update the current state queries using historical features. The updated mode and state queries are then integrated into the hybrid queries, resulting in more accurate motion forecasting and planning outcomes.}

\subsection{\ext{State anchor-based refinement}}

\ext{To fully leverage the advantages of the fine-grained representation of dynamic states, we \textit{perform state anchor-based refinement on the proposal outputs}. As shown in Figure~\ref{fig:plus} (b), the refinement process is applied independently to each trajectory across the multiple predicted modes.}

\ext{
For each trajectory, we use its state queries along with the corresponding waypoint positions. We then perform distance-aware cross-attention between these state queries and the scene context. Unlike vanilla attention, we explicitly compute the distance between each waypoint and each element in the scene context (including other agents and map features) and apply a mask to filter out distant context elements.
In this way, we perform targeted refinement for each state query using the corresponding waypoint as an anchor, and the resulting refined trajectory is then produced. The entire process operates independently for each trajectory.
}

\ext{As for the probabilities, we use the state query corresponding to the endpoint of each trajectory mode. An MLP then generates the refined probabilities, since the endpoint largely determines the overall position of the trajectory, making this approach more accurate.}

\input{figure/fig_data}

\subsection{Training objectives}

\netNameNew{} is trained in an end-to-end manner.
Specifically, for the proposal output, regression loss and classification loss are applied to supervise the accuracy of the predicted proposal trajectories and their corresponding confidence scores, collectively denoted as $\mathcal{L}_{\rm prop}$.
\ext{Subsequently, the refined output is also supervised using both regression and classification losses, which together constitute the refinement loss $\mathcal{L}_{\rm ref}$.}

We adopt the cross-entropy loss for probability score classification and the Smooth-L1 loss for trajectory regression tasks. The winner-take-all strategy is employed, optimizing only the best prediction with minimal average prediction error to the ground truth.

Additionally, we introduce two auxiliary losses, $\mathcal{L}_{\rm ts}$ and $\mathcal{L}_{\rm m}$, for intermediate features of time states and motion modes, respectively. The former enhances the coherence and causality of dynamic states across various time steps, while the latter endows the mode with distinct directional intentions. The overall loss in each sub-scene $\mathcal{L}_{\rm sub}$ is a combination of these individual losses with equal weights, formulated as:
\begin{equation}
\mathcal{L}_{\rm sub} = \mathcal{L}_{\rm prop} + \mathcal{L}_{\rm ref} + \mathcal{L}_{\rm ts} + \mathcal{L}_{\rm m}.
\end{equation}

For $\mathcal{L}_{\rm ts}$, an MLP decodes state queries into a single future trajectory $Y_{\rm ts}$, and the loss is computed against the ground truth $Y_{\rm gt}$:
\begin{equation}
\begin{split}
\mathcal{L}_{\rm ts} &= {\rm SmoothL1}(Y_{\rm ts}, Y_{\rm gt}).
\end{split}
\end{equation}

For $\mathcal{L}_{\rm m}$, MLPs decode the future trajectories $Y_{\rm m}$ and probabilities $P_{\rm m}$. 
Then the best trajectory $Y_{\rm best}$ and its corresponding probability $ P_{\rm best}$ are selected by comparing $Y_{\rm m}$ with $Y_{\rm gt}$, and the loss $\mathcal{L}_{\rm m}$ is defined as:
\begin{equation}
\begin{split}
Y_{\rm best},\, P_{\rm best} &= {\rm SelectBest}(Y_{\rm m}, Y_{\rm gt}), \\
\mathcal{L}_{\rm m} &= {\rm SmoothL1}(Y_{\rm best}, Y_{\rm gt}) + {\rm CE}(P_{\rm m}, P_{\rm best}).
\end{split}
\end{equation}

\input{figure/fig_e2e}

\ext{
For the cross-scene intention interaction, we divide the entire scene into $N_{\rm sub}$ sub-scenes and compute all losses for each sub-scene. The overall loss $\mathcal{L}$ is defined as:}
\ext{
\begin{equation}
\mathcal{L} = \mathcal{L}_{sub}^{1}+...+\mathcal{L}_{sub}^{N_{\rm sub}}.
\end{equation}
}

\section{\ext{Motion Decoupling for end-to-end planning}}

\ext{For the end-to-end planning task, the input consists of sensor data such as camera and LiDAR information, and the final planning trajectories are directly generated by a unified model. Auxiliary tasks, including detection, map segmentation, and agent motion prediction, are commonly integrated to enhance scene understanding and support safer planning.}
\ext{Next, we \textit{further extend our motion decoupling to end-to-end planning}, resulting in \textbf{\netNameetwoeNew{}} as illustrated in Figure~\ref{fig:maine2e}.}

\subsection{\ext{Scene context encoding}}
\ext{The multi-modal sensor encoder can process heterogeneous data to build a comprehensive scene representation. Specifically, multi-view images $\mathcal{I}$ and LiDAR observations $\mathcal{P}$ are fused into a bird’s-eye view (BEV) feature $F_{\rm bev} \in \mathbb{R}^{H \times W \times C}$, where $H$ and $W$ define the spatial resolution, $C$ denotes the channel dimension. 
To effectively combine visual and geometric information into a unified BEV embedding,
we follow prior methods~\cite{diffusiondrive,wote,hydraMDP} and utilize TransFuser~\cite{TransFuser}.
Agent feature $F_{\rm agent} \in \mathbb{R}^{N_{\rm agent} \times C}$ is extracted from the BEV feature, where $N_{\rm agent}$ denotes the number of surrounding agents. The BEV and agent features are then decoded with lightweight decoders
to BEV segmentation map and the positions of the surrounding agents, respectively.
While the ego status is encoded by an MLP to produce the ego feature $F_{\rm ego} \in \mathbb{R}^{1 \times C}$.}

\subsection{\ext{Trajectory decoding with decoupled queries}}
\ext{As shown in the right part of Figure~\ref{fig:maine2e}, and consistent with the practice in motion forecasting and planning tasks, we initialize two types of queries for planning: mode queries $Q_{\rm m}$ and state queries $Q_{\rm s}$. For both types of queries, cross-attention is performed with the BEV features $F_{\rm bev}$, agent features $F_{\rm agent}$, and ego features $F_{\rm ego}$. Subsequently, the self-attention and Mamba mechanisms are applied in a manner consistent with those used in \netNameNew{} for motion forecasting and planning. Similarly, the planning results generated by the mode and state queries are also output for supervision, as described above.}

\ext{After the mode queries and state queries are separately optimized, they are combined to form the hybrid motion-state queries $Q_{\rm h}$. Cross-attention is then performed with the agent features $F_{\rm agent}$ and ego features $F_{\rm ego}$. Subsequently, the self-attention and Mamba mechanisms are applied in a manner consistent with those used in \netNameNew{} for motion forecasting and planning. 
Different from \netNameNew{}, which refines the trajectories after proposal generation, \netNameetwoeNew{} directly employs a deformable attention mechanism~\cite{deformableattn} to use hybrid motion-state queries for adaptively capturing features from the BEV features $F_{\rm bev}$.
Due to the lack of sequential sensor information in NAVSIM~\cite{navsim}, the cross-scene intention interaction is excluded from \netNameetwoeNew{}. 
Finally, the hybrid queries generate the final multi-mode planning results.}

\subsection{\ext{Training objectives}}
\ext{The model is trained in an end-to-end manner, and the losses are composed of five parts. As described above, $\mathcal{L}_{\rm ts}$ and $\mathcal{L}_{\rm m}$ are derived from the planning results generated by the state queries and mode queries, respectively, while the final planning loss $\mathcal{L}_{\rm final}$ is obtained from the hybrid motion-state queries. In addition to these components, the BEV segmentation loss $\mathcal{L}_{\rm bev}$ and the surrounding agent detection loss $\mathcal{L}_{\rm agent}$ are also included, which are computed from the BEV feature and the agent feature. The overall loss $\mathcal{L}$ is a combination of these individual losses with equal weights, formulated as follows:}
\ext{
\begin{equation}
\mathcal{L} = \mathcal{L}_{\rm bev} + \mathcal{L}_{\rm agent} + \mathcal{L}_{\rm ts} + \mathcal{L}_{\rm m} + \mathcal{L}_{\rm final}.
\end{equation}
}

%% file: figure/fig_main.tex

\begin{figure*}[t!]
\centering
\includegraphics[width=1\textwidth]{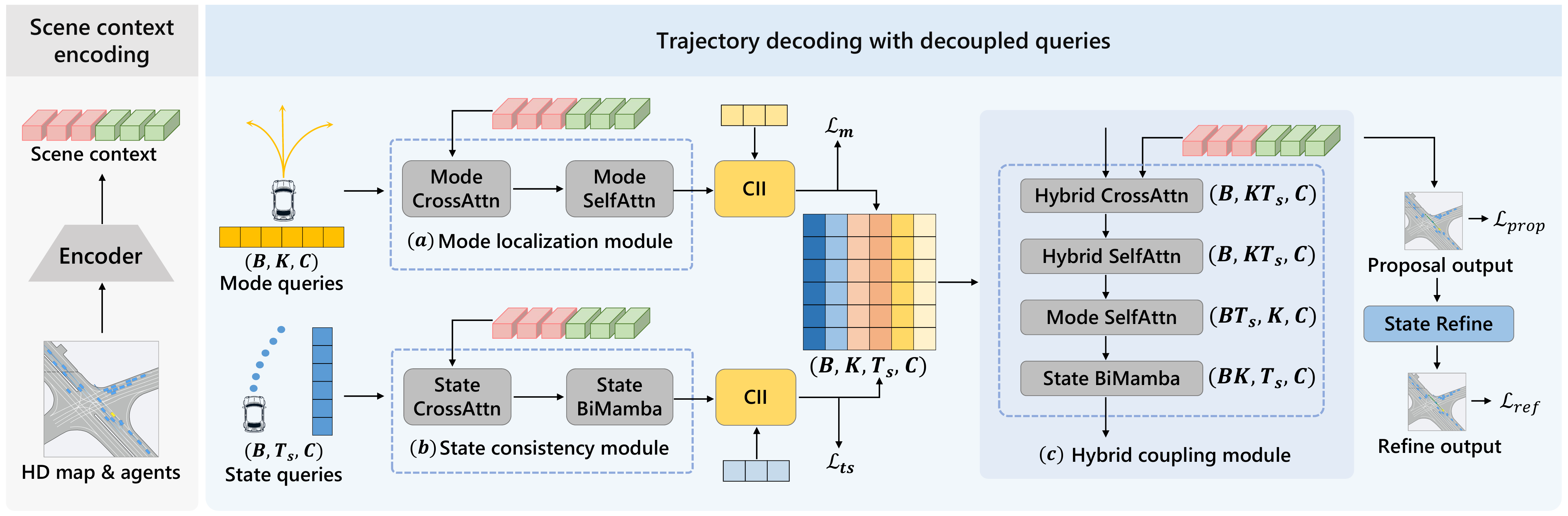}
\caption{
Overview of our \textbf{\netNameNew{}} framework: The HD maps and agents are first processed by the encoder to obtain the scene context. The decoding pipeline includes: (a) the Mode Localization Module, which processes mode queries by interacting with the scene context from the encoder and among themselves; (b) the State Consistency Module, which processes state queries; and (c) the Hybrid Coupling Module, which combines these queries to generate the final output. The feature dimension is illustrated in the figure, where $B$ represents the batch size. ``CII" indicates Cross-scene Intention Interaction with historical mode queries and state queries. ``State Refine" indicates State Anchor-based Refinement. The details of these designs are illustrated in Figure~\ref{fig:plus}.
}
\label{fig:main}
\end{figure*}

%% file: figure/fig_plus.tex

\begin{figure*}[t!]
\centering
\includegraphics[width=1\textwidth]{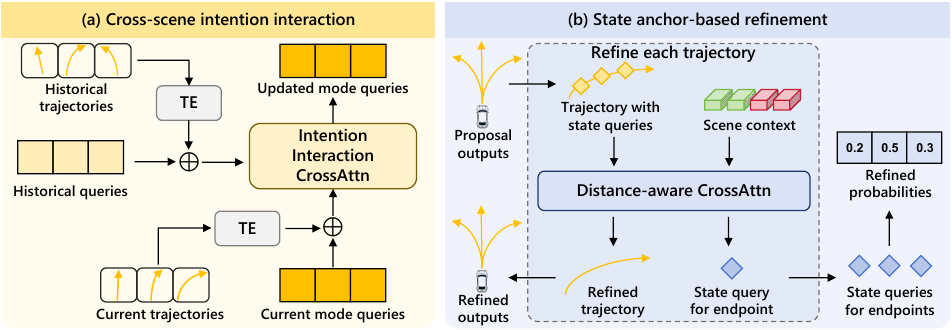}
\caption{
\ext{
(a) {\bf Cross-scene intention interaction}: the mode queries interact with historical mode queries using trajectory embeddings (TE); similarly, the state queries interact in the same manner. (b) {\bf State anchor-based refinement}: the state queries within each trajectory interact with the scene context to refine both the predicted trajectories and their associated probabilities.}
}
\label{fig:plus}
\end{figure*}

%% file: figure/fig_data.tex

\begin{figure}[t!]
\centering
\includegraphics[width=0.49\textwidth]{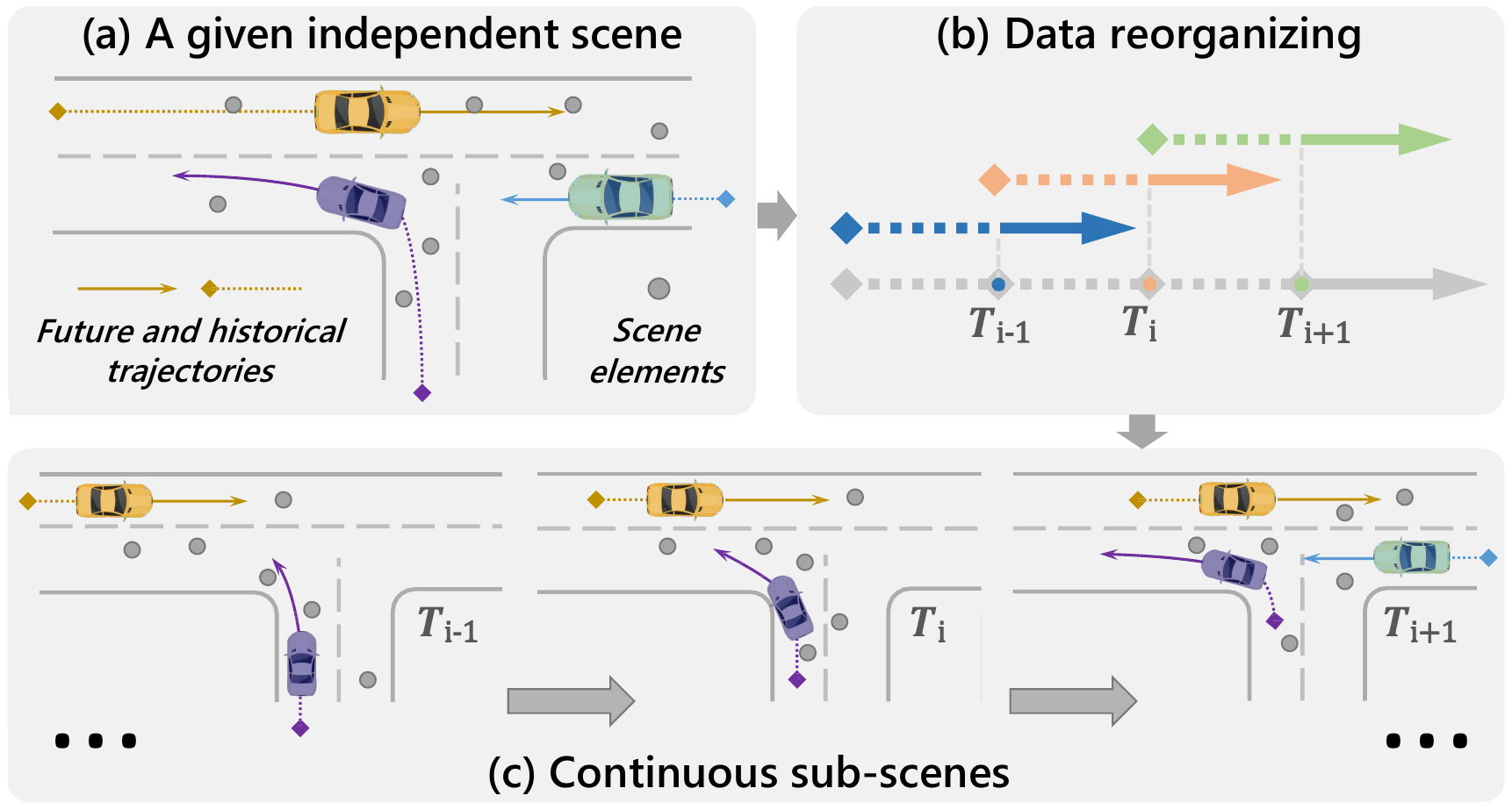}
\caption{
\ext{
Illustration of our data reorganization strategy: starting from (a) an independent scene, we (b) reorganize the trajectories into segments and aggregate surrounding elements, resulting in (c) continuous sub-scenes.
}
}
\label{fig:data}
\end{figure}

%% file: figure/fig_e2e.tex

\begin{figure*}[t!]
\centering
\includegraphics[width=1\textwidth]{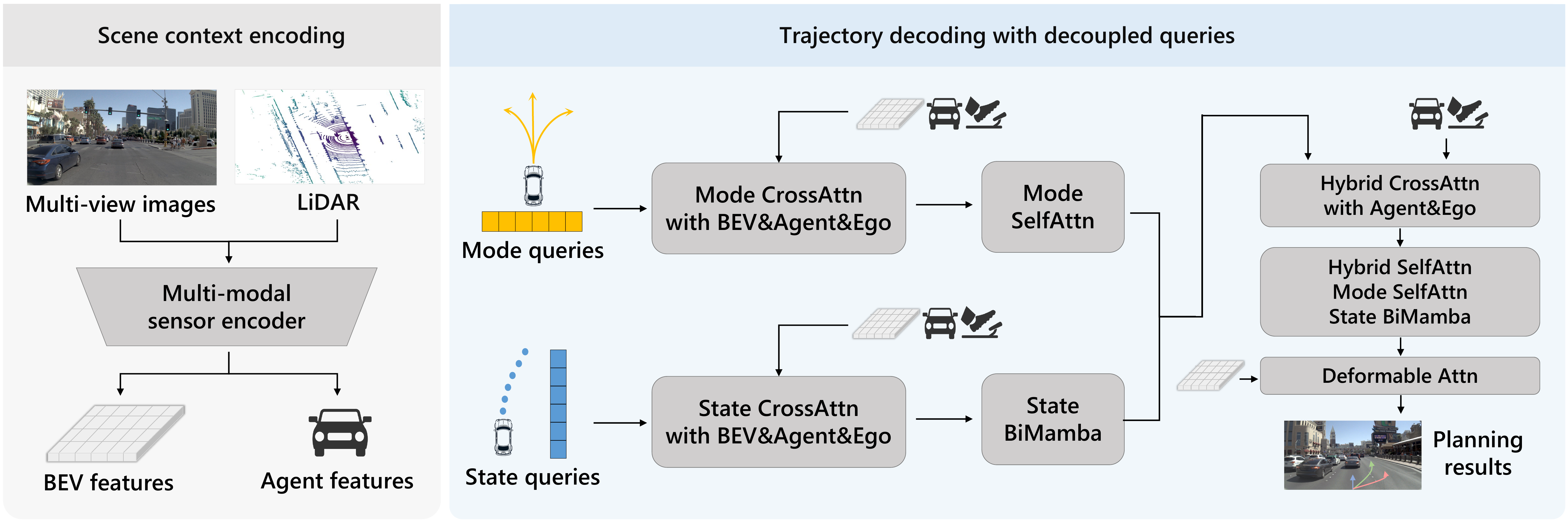}
\caption{
\ext{
Overview of our \textbf{\netNameetwoeNew{}} framework. Multi-view images and LiDAR data are first processed by a multi-modal sensor encoder to extract BEV features and agent features, which together constitute the scene context.
For trajectory decoding, two types of queries—mode queries and state queries—are initialized. Cross-attention is then performed for both query types with the BEV features, agent features, and ego vehicle status. The self-attention and Mamba mechanisms are consistent with those used in \netNameNew{} for motion forecasting and planning.
Subsequently, the mode queries and state queries are coupled to form hybrid queries, which also interact with agent features and ego status via cross-attention. Again, self-attention and Mamba mechanisms, consistent with \netNameNew{}, are applied. In addition, deformable attention is employed to adaptively extract features from the BEV representation for each state query.
Finally, the framework outputs multi-mode planning results.
}
}
\label{fig:maine2e}
\end{figure*}

%% file: section/4_experiment.tex
\section{Experiments}
\label{experiments}

\subsection{Experimental settings}

\paragraph{Datasets}

For the motion forecasting task, we evaluate the performance of our method on the Argoverse 2~\cite{argoverse2} and nuScenes~\cite{nuscenes} datasets. The Argoverse 2 dataset comprises 250,000 scenarios sampled at 10 Hz, each providing 5 seconds of historical trajectory and requiring prediction of the subsequent 6 seconds. The nuScenes dataset includes 1,000 scenes sampled at 2 Hz, with 2 seconds of past trajectory used to predict the next 6 seconds.

\ext{For the motion planning task, we evaluate our method on the nuPlan~\cite{nuplan} dataset. This large-scale closed-loop planning platform contains 1,300 hours of real-world driving data across 75 urban scenarios, providing 1 million training cases. Its simulator runs scenarios for 15 seconds at 10 Hz.}

\ext{For the end-to-end planning task, we evaluate our method on the NAVSIM~\cite{navsim} dataset. NAVSIM is a large-scale real-world autonomous driving dataset designed for non-reactive simulation and benchmarking. It integrates sensor data from eight cameras and five LiDAR sensors, together with annotated HD maps and object bounding boxes, all recorded at a frequency of 2 Hz. The dataset is divided into two subsets: navtrain, which consists of 1,192 scenarios for training and validation, and navtest, which includes 136 scenarios for testing.}

\input{table/table_av2_single}
\input{table/table_nus}

\paragraph{Evaluation metrics}
For the motion forecasting task, we adopt common metrics including minimum Average Displacement Error ($minADE_{k}$), minimum Final Displacement Error ($minFDE_{k}$), Miss Rate ($MR_{k}$), and Brier minimum Final Displacement Error ($b\text{-}minFDE_{k}$). The Argoverse 2 dataset is evaluated across 6 prediction modes, while nuScenes is evaluated across 10 prediction modes. Following the evaluation protocols of the official leaderboards, we set $K$ to 1 and 6 for the Argoverse 2 dataset, and to 5 and 10 for the nuScenes dataset.

\ext{For the motion planning task, nuPlan evaluates performance using three key metrics: the open-loop score (OLS), the non-reactive closed-loop score (NR-CLS), and the reactive closed-loop score (R-CLS). The evaluation is conducted across 6 planning modes.}

\ext{For the end-to-end planning task, the planned trajectories are evaluated using a set of closed-loop metrics, including No At-Fault Collisions ($S_{\rm NC}$), Drivable Area Compliance ($S_{\rm DAC}$), Time to Collision with bounds ($S_{\rm TTC}$), Ego Progress ($S_{\rm EP}$), Comfort ($S_{\rm CF}$), and Driving Direction Compliance ($S_{\rm DDC}$). The PDM Score ($S_{\rm PDM}$) is a composite metric derived from these individual measures, as shown below:}
\ext{
\begin{equation}
\begin{aligned}
    S_{\rm PDM} = &\ S_{\rm NC} \times S_{\rm DAC} \times \\
           &\ \left( \frac{5 \times S_{\rm EP} + 5 \times S_{\rm TTC} + 2 \times S_{\rm CF}}{12} \right).
\end{aligned}
\label{eq:pdm_scorer}
\end{equation}
}

\input{table/table_nuplan}

\paragraph{Implementation details}
\label{impl}

For the motion forecasting task, our models are trained for 60 epochs using the AdamW~\cite{adamw} optimizer with a batch size of 16 per GPU. The training is conducted with a learning rate of $3 \times 10^{-3}$ and a weight decay of $1 \times 10^{-2}$. An agent-centric coordinate system is adopted, and scene elements within a 150-meter radius of the agents of interest are sampled. The dropout rate is set to 0.2. A cosine learning rate schedule is employed, with a warm-up phase of 10 epochs.

\ext{For the motion planning task, our models are trained for 25 epochs, including a warm-up phase of 3 epochs. Training is conducted with a weight decay of $1 \times 10^{-4}$. Other training settings follow those of the motion forecasting task.}

\ext{For the end-to-end planning task, our models are trained on the navtrain split with a batch size of 16 for 100 epochs. The learning rate and weight decay are both set to $1 \times 10^{-4}$, and optimization is performed using AdamW~\cite{adamw}. For a fair comparison, the image backbone follows prior work and adopts ResNet-34~\cite{resnet}. The input consists of three images (front-right, front, and front-left), which are concatenated into a resolution of $1024 \times 256$, along with a rasterized BEV LiDAR representation. The number of planning modes is set to 20.}

All models are trained in an end-to-end manner. All experiments are conducted on eight NVIDIA GeForce RTX 3090 GPUs.  
\ext{
For \netNameNew{}, we reorganize the Argoverse 2 dataset into three continuous and evenly spaced sub-scenes, each using 3 seconds of historical data to predict the following 6 seconds.
Similarly, the nuPlan dataset is divided into two continuous and evenly spaced sub-scenes, where each sub-scene uses 1.5 seconds of history to predict the next 8 seconds.
Additionally, we refine the trajectories by leveraging the scene context within a 50-meter radius around each state query.
}

\input{table/table_navsim}

\subsection{Comparison with state of the art}

\paragraph{Motion forecasting}
We compare our methods, \netName{} and \netNameNew{}, with several existing models on the Argoverse 2~\cite{argoverse2} dataset, as shown in Table~\ref{tab:av2_single}.
To ensure a comprehensive and fair comparison, all methods are evaluated without the use of model ensembling techniques.
The results demonstrate that \netName{} significantly outperforms all previous approaches, including the state-of-the-art model QCNet~\cite{qcnet} and its post-refinement variant, SmartRefine~\cite{smartrefine}.
Specifically, our method achieves substantial improvements across all metrics, particularly in terms of $minFDE_{1}$ and $minADE_{1}$, where it outperforms QCNet by 13.02\% and 11.83\%, respectively.
\ext{With the introduction of cross-scene intention interaction and state anchor-based refinement, \netNameNew{} further improves performance and achieves results significantly better than \netName{}. In particular, for $\textit{b-minFDE}_{6}$, \netNameNew{} achieves a 0.1 reduction compared to \netName{}.}

To further demonstrate the generalization ability of our model, we also evaluate the performance of \netName{} and \netNameNew{} on the nuScenes~\cite{nuscenes} motion forecasting benchmark. The results on the test split are presented in Table~\ref{tab:nus}. Our method outperforms all other approaches across all metrics.

\paragraph{\ext{Motion planning}}
\ext{We evaluate our \netName{} and \netNameNew{} on the nuPlan~\cite{nuplan} dataset, selecting the widely used and more challenging Test 14 Hard benchmark. As shown in Table~\ref{tab:nuplan}, \netNameNew{} outperforms previous methods in terms of the open-loop score (OLS) and achieves comparable closed-loop performance to state-of-the-art approaches, including BeTopNet~\cite{betop} and DiffusionPlanner~\cite{diffusionplanner}.}

\paragraph{\ext{End-to-end planning}}
\ext{We evaluate our \netNameetwoeNew{} on the challenging NAVSIM~\cite{navsim} dataset using the navtest split. This benchmark emphasizes difficult scenarios involving dynamic intention changes while filtering out trivial cases such as stationary scenes and constant-speed driving. As shown in Table~\ref{tab:navsim}, our models outperform state-of-the-art methods, including DiffusionDrive~\cite{diffusiondrive}, WoTE~\cite{wote}, and Hydra-NeXt~\cite{hydranext}. With similar camera and LiDAR inputs and ResNet-34 used as the backbone, \netNameetwoeNew{} achieves a PDM score of 89.9, substantially surpassing all alternatives.}

\input{table/table_ab_main}

\subsection{Ablation study}

In this section, we conduct comprehensive ablation studies on \netName{} and \netNameNew{} using the validation split of the Argoverse 2~\cite{argoverse2} dataset for the motion forecasting task, in order to demonstrate the effectiveness of each model component.

\paragraph{Effects of components in \netName{} and \netNameNew{}}
Table~\ref{tab:main_abl} demonstrates the effectiveness of each component in our method. We show the baseline in the first row, which is similar to previous methods~\cite{mtr,qcnet} and utilizes mode queries to generate multi-mode future trajectories. Then, we directly adopt state queries in the second row (ID-2) to decode the trajectories. A performance decline is observed due to the surplus queries, which impose a burden on the model and make it difficult to distinguish the meanings of different types. In the third row (ID-3), we introduce two auxiliary losses, resulting in a slight improvement compared to the first row. Although the model can identify what each query represents, it demonstrates only moderate performance due to the limited information. In the fourth row (ID-4), we incorporate the three aggregation modules in Figure~\ref{fig:main} but remove auxiliary losses, leading to significant performance enhancements. In the fifth row (ID-5), our \netName{} integrates all these techniques and achieves outstanding performance.

\ext{Next, we conduct an ablation study on the components newly introduced in \netNameNew{}, namely the cross-scene intention interaction and the state anchor-based refinement. As shown in the last three rows (ID-6 to ID-8) of the table, each component contributes meaningfully to improving the model’s overall performance.}

\paragraph{Effects of state sequence modeling with Mamba in the decoder}
Mamba excels at sequence modeling, so we utilize Bidirectional Mamba~\cite{videomamba,vim} to enhance the consistency of states across different time steps. To demonstrate its effectiveness, we compare Bidirectional Mamba with several other modules, including Unidirectional Mamba~\cite{mamba}, Attention, Conv1d, and GRU~\cite{gru}. As illustrated in  Table~\ref{tab:abl_mamba_layer}, our Bidirectional Mamba configuration outperforms the others due to its specialized design for sequence modeling, compared to Attention, and its capability to perform both forward and backward scans, unlike Unidirectional Mamba.

\input{table/table_ab_mamba}

\paragraph{Effects of auxiliary losses and aggregation modules in the decoder}
We conduct an ablation study to assess the impacts of auxiliary losses and aggregation modules. As illustrated in Table~\ref{tab:abl_agg_aux}, removing any of these losses or modules leads to a performance decline in the model. Notably, the aggregation modules have a greater impact than the auxiliary losses. This is attributed to the critical role of learning information from the scene context and from each other, which is essential for decoupling queries to represent distinct meanings.

\input{table/table_ab_agg_aux}

\paragraph{Effects of state queries}
We conduct an ablation study on the number of state queries, as shown in Table~\ref{tab:sq_abl}. In our default setting, we use 60 state queries to represent the future states at 60 timestamps. As we gradually reduce the number of state queries, we observe a performance decline due to the increasing ambiguity of the state query meanings.

\input{table/table_ab_state_query}

\paragraph{Effects of the depth of Attention and Mamba blocks in the decoder}
A suitable depth configuration of Attention and Mamba units is crucial for achieving an optimal balance between efficiency and performance. As depicted in Table~\ref{tab:layer_abl}, we conduct an ablation study focusing on the layer depth. It is observed that the best results are obtained with Attention units at a depth of three and Mamba units at a depth of two.

\input{table/table_ab_layer}

\paragraph{Effects of the depth of Mamba blocks in the encoder}
We add ablation studies on the Mamba for encoding agent historical information in the encoder of our model. Table~\ref{tab:abl_seq} shows different modules for encoding the historical information of agents. Our goal is to aggregate historical information up to the present time, making Unidirectional Mamba the most suitable choice. Table~\ref{tab:abl_mamba_layer_num} presents an ablation study concerning the number of Mamba blocks, indicating that three layers yield the optimal performance.

\input{table/table_ab_seq}
\input{table/table_ab_mamba_layer}

\input{figure/fig_visual}

\subsection{An analysis to improve the measurement of motion decoupling strategy}
To thoroughly demonstrate the effectiveness of the motion decoupling strategy, we evaluate the outputs of both state queries and mode queries using $minADE$ and $minFDE$, as shown in Table~\ref{tab:analysis}. We can see that the $minADE_1$ and $minFDE_1$ of the trajectories from state query outputs are better than those from mode query outputs. This means state dynamics are encoded in state queries. Additionally, there are six output trajectories from mode queries, indicating that directional information is predominantly stored in mode queries. The final outputs take advantage of the strengths of both. 

\input{table/table_analysis}

\subsection{Efficiency analysis}
Balancing performance, inference speed, and model size is crucial for model deployment. We compare our~\netName~with two recent representative models: the state-of-the-art QCNet~\cite{qcnet} and its enhancement via post-refinement, SmartRefine~\cite{smartrefine}. Our model size is 5.9M, compared to 7.7M for QCNet and 8.0M for SmartRefine. Despite being smaller, our model significantly outperforms them, as detailed in Table~\ref{tab:av2_single}. 

Regarding inference speed, we compare~\netName~and QCNet, both end-to-end methods. The measurements are performed on the Argoverse 2 single-agent validation set using an NVIDIA GeForce RTX 3090 GPU, with a batch size of one. The average inference time of ~\netName~is only 38 ms, approximately 2.5 times faster than QCNet's 94 ms. This demonstrates that our method is not only superior in performance but also more efficient.

To provide a more comprehensive evaluation, we further compare the computational costs of recent representative methods. Table~\ref{tab:cost} provides this comparison. The experiments are conducted on the Argoverse 2~\cite{argoverse2} dataset using 8 NVIDIA GeForce RTX 3090 GPUs.

\input{table/table_cost}

\input{figure/fig_visuale2e}

\subsection{Qualitative results}
In Figure~\ref{fig:visual}, we present qualitative results of our network for the motion forecasting task on the Argoverse 2 dataset in the validation split. The results of the baseline model, which lacks the decoupled query paradigm, are shown in panel (a), while the results of our~\netName~are shown in panel (b). 
From the first two rows, it is evident that by explicitly optimizing the dynamic states of future trajectories, our model predicts trajectories that are more accurate and closer to the ground truth. From the third row, it is apparent that our model can better capture potential directional intentions.

\ext{In Figure~\ref{fig:visual_e2e}, we present qualitative results of \netNameetwoeNew{} for the end-to-end planning task on the NAVSIM dataset. The results demonstrate that our model generates accurate plans in both straight-driving and left-turn scenarios.}

\input{figure/fig_fail}

\subsection{Failure cases}
Although our \netName{} demonstrates exceptional performance, it still has failure cases. We analyze these typical examples and present qualitative results to illustrate scenarios where the model underperforms, as shown in Figure~\ref{fig:visual_fail}. This analysis aims to guide future efforts toward developing more robust and reliable algorithms.
In the first row, the vehicle intends to turn into an alley, reflecting subjective driving behavior. However, the model predicts that it will continue straight. Improving predictions in such cases may require incorporating additional cues about driver intent, such as turn signals.
In the second row, the agent must navigate through a complex intersection to reach one of several roads, but the model fails to capture this behavior accurately. This inaccuracy may stem from an incomplete understanding of the complex map topology and the unbalanced distribution of driving data. Addressing data imbalance is essential to resolve this issue.

%% file: table/table_av2_single.tex

\begin{table*} [t!]
\caption{Performance of motion forecasting \textit{on the Argoverse 2 dataset in the test split}. For each metric, the best result is in \textbf{bold} and the second best result is \underline{underlined}. All the results are obtained using individual models without ensembling.
}
\label{tab:av2_single}
\centering
\begin{tabular}{l|c|cccccc}

\toprule[1.5pt]
Method & Reference & $\textit{minFDE}_{1}$ $\downarrow$ & $\textit{minADE}_{1}$ $\downarrow$ & $\textit{minFDE}_{6}$ $\downarrow$ & $\textit{minADE}_{6}$ $\downarrow$ & $\textit{MR}_{6}$ $\downarrow$ & $\textit{b-minFDE}_{6}$ $\downarrow$\\
[1.5pt]\hline\noalign{\vskip 2pt}
FRM~\cite{frm} & ICLR 2023                      & 5.93 & 2.37 & 1.81 & 0.89 & 0.29 & 2.47\\
HDGT~\cite{hdgt} & TPAMI 2023                   & 5.37 & 2.08 & 1.60 & 0.84 & 0.21 & 2.24\\  
SIMPL~\cite{simpl} & RA-L 2024                  & 5.50 & 2.03 & 1.43 & 0.72 & 0.19 & 2.05\\
THOMAS~\cite{thomas} & ICLR 2022                & 4.71 & 1.95 & 1.51 & 0.88 & 0.20 & 2.16\\
GoRela~\cite{gorela} & ICRA 2023                & 4.62 & 1.82 & 1.48 & 0.76 & 0.22 & 2.01\\ 
MTR~\cite{mtr} & NeurIPS 2022                   & 4.39 & 1.74 & 1.44 & 0.73 & 0.15 & 1.98\\
HPTR~\cite{HPTR} & NeurIPS 2023                 & 4.61 & 1.84 & 1.43 & 0.73 & 0.19 & 2.03\\
GANet~\cite{ganet} & ICRA 2023                  & 4.48 & 1.77 & 1.34 & 0.72 & 0.17 & 1.96\\
ProphNet~\cite{prophnet} & CVPR 2023            & 4.74 & 1.80 & 1.33 & 0.68 & 0.18 & 1.88\\  
QCNet~\cite{qcnet} & CVPR 2023                  & 4.30 & 1.69 & 1.29 & 0.65 & 0.16 & 1.91\\   
CaDeT~\cite{cadet} & CVPR 2024                  & 4.33 & 1.74 & 1.24 & 0.67 & 0.15 & 1.86\\
RealMotion~\cite{RealMotion} & NeurIPS 2024     & 3.93 & 1.59 & 1.24 & 0.66 & 0.15 & 1.89 \\
SmartRefine~\cite{smartrefine} & CVPR 2024      & 4.17 & 1.65 & 1.23 & \underline{0.63} & 0.15 & 1.86\\
[1.5pt] \hline\noalign{\vskip 2pt}
\bf\netName~ & Ours               & \underline{3.74} &  \bf1.49 & \underline{1.17} & \bf0.61 &\underline{0.13} & \underline{1.84} \\
\ext{\bf\netNameNew} & \ext{Ours} & \ext{\bf3.70} & \ext{\underline{1.50}} & \ext{\bf1.12} & \ext{\bf0.61} & \ext{\bf0.12} & \ext{\bf1.74} \\

\bottomrule[1.5pt]
\end{tabular}
\end{table*}

%% file: table/table_nus.tex

\begin{table} [tb]
\caption{Performance of motion forecasting \textit{on the nuScenes dataset in the test split}. ``-'': Unknown.}
\label{tab:nus}
\centering
\setlength{\tabcolsep}{1.7mm}
{\begin{tabular}{l|ccccc}
\toprule[1.5pt]
Method & $\textit{minFDE}_{1}$ & $\textit{minADE}_{5}$ & $\textit{minADE}_{10}$ & $\textit{MR}_{5}$ & $\textit{MR}_{10}$ \\ 
[1.5pt]\hline\noalign{\vskip 2pt}

Trajectron++~\cite{trajectron++}     & 9.52 &  1.88 & 1.51 & 0.70 & 0.57 \\  
LaPred~\cite{lapred}                 & 8.37 &  1.47 & 1.12 & 0.53 & 0.46 \\
P2T~\cite{p2t}                       & 10.50 &  1.45 & 1.16 & 0.64 & 0.46 \\
GOHOME~\cite{gohome}                 & 6.99 &  1.42 & 1.15 & 0.57 & 0.47 \\
CASPNet~\cite{caspnet}               & - &  1.41 & 1.19 & 0.60 & 0.43 \\
Autobot~\cite{Autobot}               & 8.19 &  1.37 & 1.03 & 0.62 & 0.44 \\
THOMAS~\cite{thomas}                 & 6.71 &  1.33 & 1.04 & 0.55 & 0.42 \\
PGP~\cite{pgp}                       & 7.17 &  1.27 & 0.94 & 0.52 & \underline{0.34} \\
LAformer~\cite{laformer}             & 6.95 &  \underline{1.19} & 1.19 & 0.48 & 0.48 \\
[1.5pt]\hline\noalign{\vskip 2pt}

\bf\netName~(Ours)                         & \underline{6.60} & 1.22 & \underline{0.89} & \underline{0.43} & \underline{0.34}  \\
\ext{\bf\netNameNew~(Ours)}                & \ext{\bf6.33} &  \ext{\bf1.18} & \ext{\bf0.87} & \ext{\bf0.40} & \ext{\bf0.33}  \\

\bottomrule[1.5pt]
\end{tabular}}
\end{table}

%% file: table/table_nuplan.tex

\begin{table} [t!]
\caption{\ext{Performance of open-loop and closed-loop motion planning \textit{on the nuPlan dataset in the Test 14 Hard split}.}}
\label{tab:nuplan}
\centering
\setlength{\tabcolsep}{2.2mm}
{
\begin{tabular}{l|l|ccc}
\toprule[1.5pt]

\ext{Paradigm} & \ext{Method} & \ext{OLS $\uparrow$} & \ext{NR-CLS $\uparrow$} & \ext{R-CLS $\uparrow$} \\
[1.5pt]\hline\noalign{\vskip 2pt}

\multirow{2}{*}{\ext{Rule}} & \ext{IDM}~\cite{idm} & \ext{0.20} & \ext{0.56} & \ext{0.62} \\
                            & \ext{PDM-Closed}~\cite{PDM} & \ext{0.26} & \ext{0.65} & \ext{0.75} \\
\midrule
\multirow{2}{*}{\ext{Hybrid}} & \ext{GameFormer}~\cite{gameformer} & \ext{0.75} & \ext{0.67} & \ext{0.69} \\
                              & \ext{PDM-Hybrid}~\cite{PDM} & \ext{0.74} & \ext{0.66} & \ext{0.76} \\
\midrule
\multirow{9}{*}{\ext{Learning}} & \ext{UrbanDriver}~\cite{UrbanDriver} & \ext{0.77} & \ext{0.52} & \ext{0.49} \\
                                & \ext{PDM-Open}~\cite{PDM} & \ext{0.79} & \ext{0.34} & \ext{0.36} \\   
                                & \ext{PlanCNN}~\cite{plant} & \ext{0.52} & \ext{0.49} & \ext{0.52} \\
                                & \ext{GC-PGP}~\cite{GCPGP} & \ext{0.74} & \ext{0.43} & \ext{0.40} \\
                                & \ext{PlanTF}~\cite{plantf} & \ext{0.83} & \ext{0.73} & \ext{0.62} \\
                                & \ext{BeTopNet}~\cite{betop} & \ext{0.84} & \bf \ext{0.77} & \bf \ext{0.69} \\
                                & \ext{DiffusionPlanner}~\cite{diffusionplanner} & \ext{-} & \ext{\underline{0.76}} & \bf \ext{0.69} \\
                                [1.5pt]\cline{2-5}\noalign{\vskip 2pt}
                                & \bf \ext{\netName{}~(Ours)} & \ext{\underline{0.86}} & \ext{0.73} & \ext{\underline{0.67}} \\
                                & \bf \ext{\netNameNew~(Ours)} & \bf \ext{0.88} & \ext{\underline{0.76}} & \bf \ext{0.69} \\

\bottomrule[1.5pt]
\end{tabular}
}


\end{table}

%% file: table/table_navsim.tex
\begin{table*} [t!]
\centering
\caption{
\ext{Performance of end-to-end planning \textit{on the NAVSIM dataset in the navtest split} under the closed-loop metrics. 
``C": Camera, ``L": LiDAR;
The backbone of all methods is consistently ResNet-34.}
}
\label{tab:navsim}

{\begin{tabular}{l|c|c|cccccc}
\toprule[1.5pt]
\ext{Method} & \ext{Reference} & \ext{Input} & \ext{NC $\uparrow$} & \ext{DAC $\uparrow$} & \ext{TTC $\uparrow$} & \ext{Comf. $\uparrow$} & \ext{EP $\uparrow$} & \ext{PDM Score $\uparrow$} \\
\midrule
\ext{UniAD~\cite{UniAD}}                              & \ext{CVPR 2023} & \ext{C} & \ext{97.8} & \ext{91.9} & \ext{92.9} & \ext{\textbf{100}} & \ext{78.8} & \ext{83.4} \\
\ext{LTF~\cite{TransFuser}}                           & \ext{TPAMI 2022} & \ext{C} & \ext{97.4} & \ext{92.8} & \ext{92.4} & \ext{\textbf{100}} & \ext{79.0} & \ext{83.8} \\
\ext{PARA-Drive~\cite{paradrive}}                     & \ext{CVPR 2024} & \ext{C} & \ext{97.9} & \ext{92.4} & \ext{93.0} & \ext{99.8} & \ext{79.3} & \ext{84.0} \\
\ext{LAW~\cite{LAW}}                                  & \ext{ICLR 2025} & \ext{C} & \ext{96.4} & \ext{95.4} & \ext{88.7} & \ext{\underline{99.9}} & \ext{81.7} & \ext{84.6} \\
\ext{Hydra-MDP++~\cite{HydraMDPplus}}                 & \ext{arXiv 2025} & \ext{C} & \ext{97.6} & \ext{96.0} & \ext{93.1} & \ext{\textbf{100}} & \ext{80.4} & \ext{86.6} \\
\ext{VADv2-$\mathcal{V}_{8192}$~\cite{vadv2}}         & \ext{arXiv 2024} & \ext{C \& L} & \ext{97.2} & \ext{89.1} & \ext{91.6} & \ext{\textbf{100}} & \ext{76.0} & \ext{80.9} \\
\ext{Hydra-MDP-$\mathcal{V}_{8192}$~\cite{hydraMDP}}  & \ext{arXiv 2024} & \ext{C \& L} & \ext{97.9} & \ext{91.7} & \ext{92.9} & \ext{\textbf{100}} & \ext{77.6} & \ext{83.0} \\
\ext{TransFuser~\cite{TransFuser}}                    & \ext{TPAMI 2022} & \ext{C \& L} & \ext{97.7} & \ext{92.8} & \ext{92.8} & \ext{\textbf{100}} & \ext{79.2} & \ext{84.0} \\
\ext{DRAMA~\cite{drama}}                              & \ext{arXiv 2024} & \ext{C \& L} & \ext{98.0} & \ext{93.1} & \ext{94.8} & \ext{\textbf{100}} & \ext{80.1} & \ext{85.5} \\
\ext{DiffusionDrive~\cite{diffusiondrive}}            & \ext{CVPR 2025} & \ext{C \& L} & \ext{98.2} & \ext{96.2} & \ext{94.7} & \ext{\textbf{100}} & \ext{\underline{82.2}} & \ext{88.1} \\
\ext{WoTE~\cite{wote}}                                & \ext{ICCV 2025} & \ext{C \& L} & \ext{\textbf{98.5}} & \ext{96.8} & \ext{\underline{94.9}} & \ext{\underline{99.9}} & \ext{81.9} & \ext{88.3} \\
\ext{Hydra-NeXt~\cite{hydranext}}                     & \ext{arXiv 2025} & \ext{C \& L} & \ext{98.1} & \ext{\underline{97.7}} & \ext{94.6} & \ext{\textbf{100}} & \ext{81.8} & \ext{\underline{88.6}} \\\midrule

\ext{\bf\netNameetwoeNew}                       & \ext{Ours} & \ext{C \& L} & \ext{\underline{98.4}} & \ext{\textbf{97.9}} & \ext{\bf 95.1} & \ext{\textbf{100}} & \ext{\bf 84.2} & \ext{\bf 89.9} \\

\bottomrule[1.5pt]
\end{tabular}}

\end{table*}

%% file: table/table_ab_main.tex

\begin{table*} [ht!]
\caption{Ablation study on the core components of \netNameNew{} \textit{on the Argoverse 2 dataset in the validation split}. ``Decouple Query'' indicates decoupled query paradigm. ``Agg. Module'' indicates three aggregation modules. ``Aux. Loss'' indicates two auxiliary losses. ``CII" indicates cross-scene intention interaction, and ``Refine" indicates state anchor-based refinement.}
\label{tab:main_abl}
\centering
\begin{tabular}{c|cccccc|cccccc}
\toprule[1.5pt]
\multirow{2}{*}{ID} & State & Decouple & Agg. & Aux. & \multirow{2}{*}{\ext{CII}} & \multirow{2}{*}{\ext{Refine}} & \multirow{2}{*}{$\textit{minFDE}_{1}$} & \multirow{2}{*}{$\textit{minADE}_{1}$} & \multirow{2}{*}{$\textit{minFDE}_{6}$} & \multirow{2}{*}{$\textit{minADE}_{6}$} & \multirow{2}{*}{$\textit{MR}_{6}$} & \multirow{2}{*}{$\textit{b-minFDE}_{6}$}\\
& Query & Query & Module & Loss & & & & & &\\
[1.5pt]\hline\noalign{\vskip 2pt}

1 &  &  &  &  & &  & 4.489 & 1.792 & 1.414 & 0.750 & 0.184 & 2.067\\ 
2 & \checkmark  &  &  &  & &  & 4.494 & 1.800 & 1.505 & 0.777 & 0.208 & 2.138\\
3 & \checkmark  & \checkmark &  & \checkmark & &  & 4.385 & 1.746 & 1.405 & 0.761 & 0.180 & 2.051\\
4 & \checkmark  & \checkmark & \checkmark &  & &  & 4.247 & 1.695 & 1.319 & 0.687 & 0.166 & 1.961\\
5 & \checkmark  & \checkmark & \checkmark & \checkmark & &  & 3.917 & 1.609 & 1.268 & 0.674 & 0.152 & 1.918\\
\ext{6} & \ext{\checkmark} & \ext{\checkmark} & \ext{\checkmark} & \ext{\checkmark} & \ext{\checkmark} & \ext{} & \ext{3.839} & \ext{1.550} & \ext{1.204} & \ext{0.637} & \ext{0.139} & \ext{1.832} \\
\ext{7} & \ext{\checkmark} & \ext{\checkmark} & \ext{\checkmark} & \ext{\checkmark} & \ext{} & \ext{\checkmark} & \ext{3.856} & \ext{1.568} & \ext{1.230} & \ext{0.644} & \ext{0.148} & \ext{1.856} \\
\ext{8} & \ext{\checkmark} & \ext{\checkmark} & \ext{\checkmark} & \ext{\checkmark} & \ext{\checkmark} & \ext{\checkmark} & \ext{\bf 3.795} & \ext{\bf 1.533} & \ext{\bf 1.167} & \ext{\bf 0.626} & \ext{\bf 0.132} & \ext{\bf 1.794} \\

\bottomrule[1.5pt]
\end{tabular}

\end{table*}

%% file: table/table_ab_mamba.tex

\begin{table}[ht!]
\centering
\caption{Ablation study on the sequence modeling choices in the decoder. ``Uni-Mamba'' and ``Bi-Mamba'' represent Unidirectional Mamba and Bidirectional Mamba.}
\label{tab:abl_mamba_layer}
\begin{tabular}{l|ccc}
\toprule[1.5pt]
& $\textit{minFDE}_{6}$ & $\textit{minADE}_{6}$ & $\textit{MR}_{6}$\\[1.5pt]\hline\noalign{\vskip 2pt}
None    & 1.307 & 0.692 & 0.161\\
GRU     & 1.842 & 0.923 & 0.274\\
Conv1d  & 1.304 & 0.693 & 0.161\\
Attention    & 1.289 & 0.687 & 0.159\\
Uni-Mamba  & 1.288 & 0.690 & 0.156\\
Bi-Mamba   & \bf 1.268 & \bf 0.674 & \bf 0.152\\
\bottomrule[1.5pt]    
\end{tabular}

\end{table}

%% file: table/table_ab_agg_aux.tex

\begin{table}[ht!]
\centering
\caption{Ablation study on the effects of aggregation modules and auxiliary losses in the decoder. ``H.C.'' indicates Hybrid Coupling Module. ``S.C.'' indicates State Consistency Module. ``M.L.'' indicates Mode Localization Module.}

\begin{tabular}{l|ccc}
\toprule[1.5pt]
& $\textit{minFDE}_{6}$ & $\textit{minADE}_{6}$ & $\textit{MR}_{6}$\\[1.5pt]\hline\noalign{\vskip 2pt}
Without $\mathcal{L}_{\rm ts}$     & 1.290 & 0.715 & 0.161\\
Without $\mathcal{L}_{\rm m}$      & 1.289 & 0.687 & 0.159\\
Without H.C.                       & 1.324 & 0.704 & 0.164\\
Without S.C.                       & 1.317 & 0.697 & 0.162\\
Without M.L.                       & 1.297 & 0.693 & 0.158\\
All       & \bf1.268 & \bf0.674 & \bf0.152\\
\bottomrule[1.5pt]
\end{tabular}
\label{tab:abl_agg_aux}
\end{table}

%% file: table/table_ab_state_query.tex

\begin{table}[ht!]
\centering
\caption{Ablation study on the number of state queries.}
\label{tab:sq_abl}
\begin{tabular}{c|ccc}
\toprule[1.5pt]
Queries & $\textit{minFDE}_{6}$ & $\textit{minADE}_{6}$ & $\textit{MR}_{6}$\\[1.5pt]\hline\noalign{\vskip 2pt}
10     & 1.312 & 0.704 & 0.160\\
20     & 1.294 & 0.688 & 0.157\\
30     & 1.290 & 0.692 & 0.155\\
60     & \bf 1.268 & \bf 0.674 & \bf 0.152\\
\bottomrule[1.5pt]
\end{tabular}

\end{table}

%% file: table/table_ab_layer.tex

\begin{table}[ht!]
\centering
\caption{Ablation study on the depth of Attention and Mamba layers in the decoder.}
\label{tab:layer_abl}

\begin{tabular}{c|c|ccc}
\toprule[1.5pt]        
Attention &  Mamba & $\textit{minFDE}_{6}$ & $\textit{minADE}_{6}$ & $\textit{MR}_{6}$\\
[1.5pt]\hline\noalign{\vskip 1pt}
1 & 1 & 1.309 & 0.708 & 0.160\\
\hline\noalign{\vskip 1pt}
2 & \multirow{2}{*}{2} & 1.288 & 0.691 & 0.157\\
\cline{1-1}\cline{3-5}\noalign{\vskip 1pt}
\multirow{2}{*}{3} &  & \bf 1.268 & \bf 0.674 & \bf 0.152\\
\cline{2-5}\noalign{\vskip 1pt}
& 3 & 1.276 & 0.675 & 0.154\\
\bottomrule[1.5pt]
\end{tabular}

\end{table}

%% file: table/table_ab_seq.tex

\begin{table}[ht!]
\centering
\caption{Ablation study on the sequence modeling choices in the encoder.}
\begin{tabular}{l|ccc}
\toprule[1.5pt]
& $\textit{minFDE}_{6}$ & $\textit{minADE}_{6}$ & $\textit{MR}_{6}$\\[1.5pt]\hline\noalign{\vskip 2pt}
GRU     & 1.344 & 0.726 & 0.170\\
Bi-Mamba  & 1.280 & 0.684 & 0.154\\
Uni-Mamba   & \bf 1.268 & \bf 0.674 & \bf 0.152\\
\bottomrule[1.5pt]
\end{tabular}
\label{tab:abl_seq}
\end{table}

%% file: table/table_ab_mamba_layer.tex

\begin{table}[ht!]
\centering
\caption{Ablation study on the depth of Mamba blocks in the encoder.}
\begin{tabular}{c|ccc}
\toprule[1.5pt]
Number & $\textit{minFDE}_{6}$ & $\textit{minADE}_{6}$ & $\textit{MR}_{6}$\\[1.5pt]\hline\noalign{\vskip 2pt}
1         & 1.312 & 0.701 & 0.162\\
2         & 1.283 & 0.681 & 0.155\\
3         & \bf 1.268 & \bf 0.674 & \bf 0.152\\
\bottomrule[1.5pt]
\end{tabular}
\label{tab:abl_mamba_layer_num}
\end{table}

%% file: figure/fig_visual.tex

\begin{figure*}[t!]
\centering
\includegraphics[width=1\textwidth]{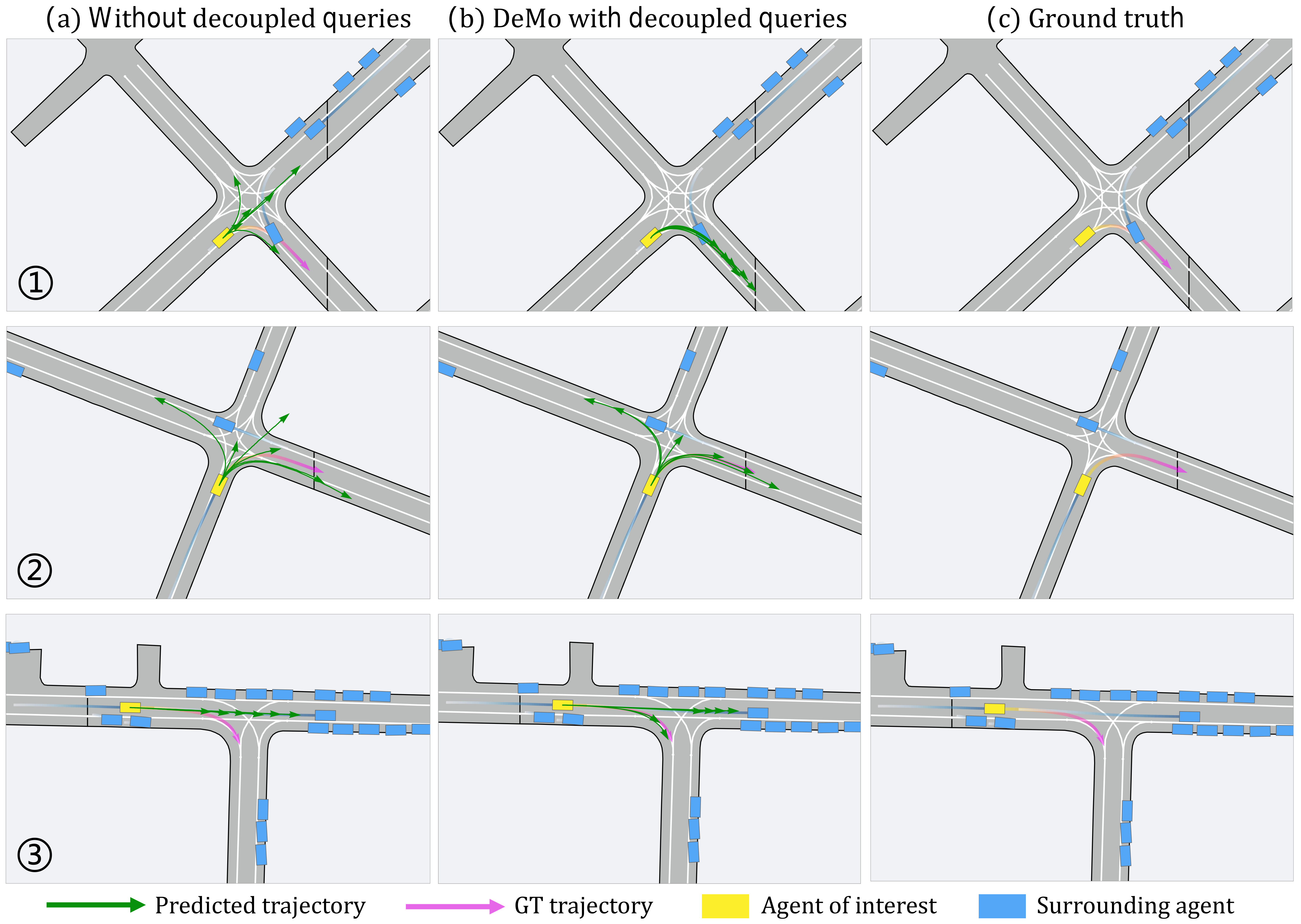}
\caption{
Qualitative results for the motion forecasting task \textit{on the Argoverse 2 dataset in the validation split}. Panel (a) illustrates the results of the baseline model without decoupled queries; Panel (b) illustrates the results of our \netName{}, which employs decoupled queries; and Panel (c) represents the ground truth.
}
\label{fig:visual}
\end{figure*}

%% file: table/table_analysis.tex

\begin{table} [ht!]
\caption{An analysis to improve the measurement of motion decoupling strategy.}
\label{tab:analysis}
\centering
{\begin{tabular}{l|cccc}
\toprule[1.5pt]
$\textbf{}$ & $\textit{minFDE}_{1}$ & $\textit{minADE}_{1}$ & $\textit{minFDE}_{6}$ & $\textit{minADE}_{6}$ \\
[1.5pt]\hline\noalign{\vskip 2pt}
State query out   & 3.84 & 1.52 & - & - \\   
Mode query out   & 4.12 & 1.63 & 1.31 & 0.67 \\
Final out   & 3.93 & 1.54 & 1.24 & 0.64 \\
\bottomrule[1.5pt]
\end{tabular}}
\end{table}

%% file: table/table_cost.tex

\begin{table} [ht!]
\caption{Comparison of computational cost with other recent representative methods. ``BS'' indicates batch size. ``Train'' indicates training time.}
\label{tab:cost}
\centering
{\begin{tabular}{l|ccccc}
\toprule[1.5pt]
$\textbf{Method}$ & FLOPs & Train & Memory & Parameter & BS\\[1.5pt]\hline\noalign{\vskip 2pt}
SIMPL~\cite{simpl} & 19.7 GFLOPs & 8h & 14G & 1.9M & 16\\
QCNet~\cite{qcnet} & 53.4 GFLOPs & 45h & 16G & 7.7M & 4\\[1.5pt]
\hline\noalign{\vskip 2pt}
\bf\netName~(Ours) & 22.8 GFLOPs & 9h & 12G & 5.9M & 16\\
\bottomrule[1.5pt]
\end{tabular}}

\end{table}

%% file: figure/fig_visuale2e.tex

\begin{figure}[t!]
\centering
\includegraphics[width=0.48\textwidth]{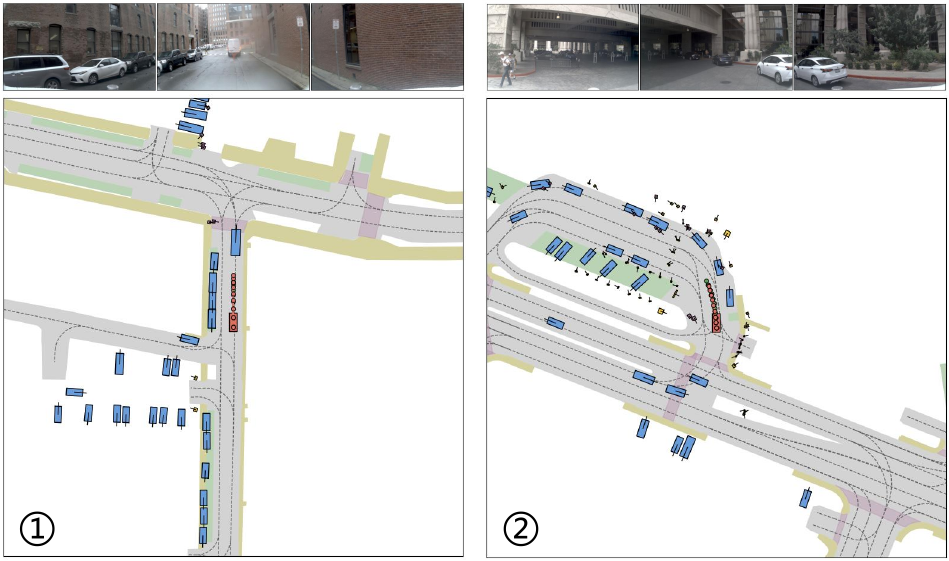}
\caption{
\ext{Qualitative results for end-to-end planning \textit{on the NAVSIM dataset}. The visualization includes three front-facing camera views: front-left, front, and front-right. The trajectory planned by \netNameetwoeNew{} is shown in orange, while the ground-truth trajectory is shown in green.}
}
\label{fig:visual_e2e}
\end{figure}

%% file: figure/fig_fail.tex

\begin{figure}[t!]
\centering
\includegraphics[width=0.49\textwidth]{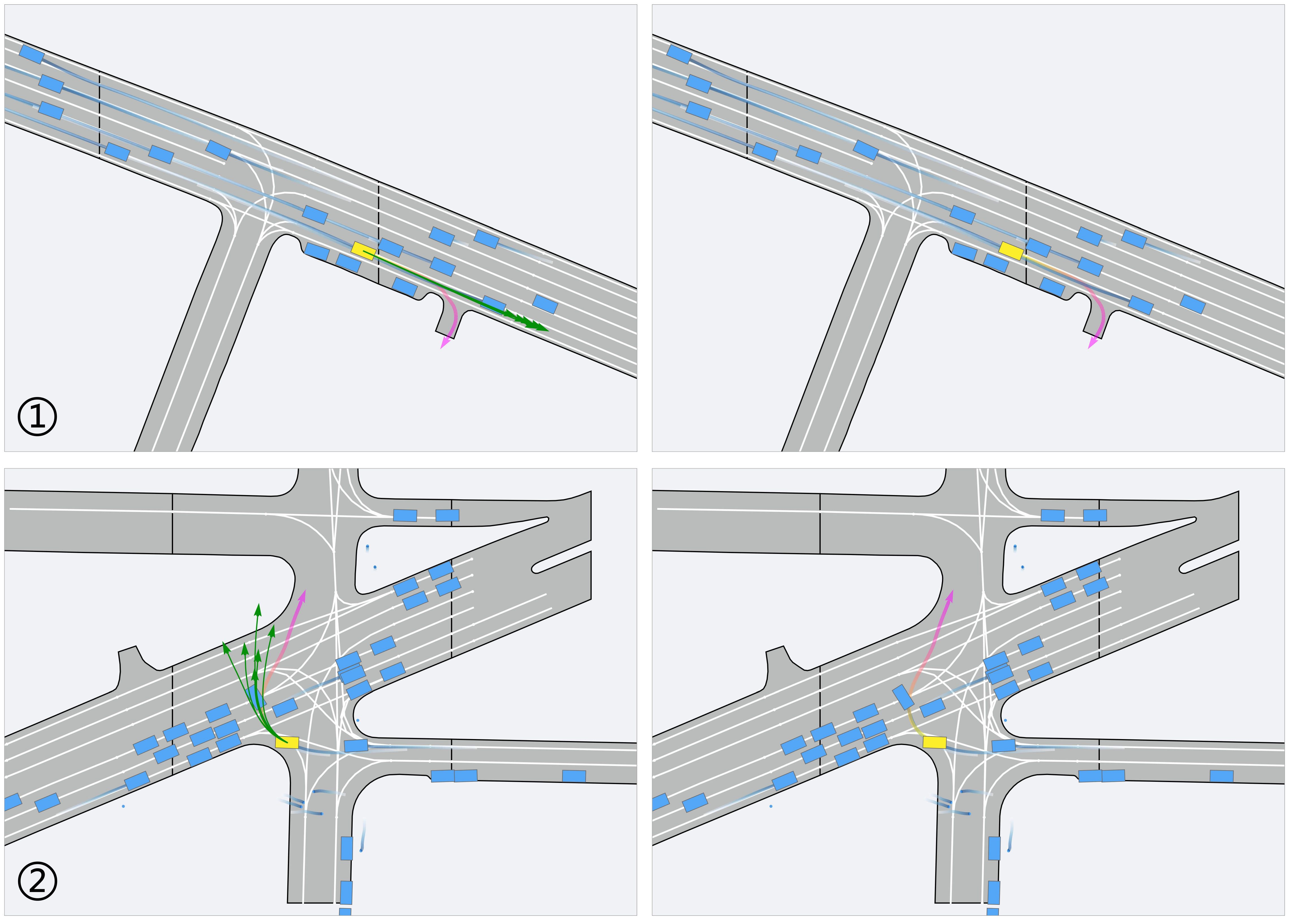}
\caption{
Failure cases \textit{on the Argoverse 2 dataset in the validation split}. The left panel shows our model’s predictions, while the right panel shows the ground-truth trajectories.
}
\label{fig:visual_fail}
\end{figure}

%% file: section/5_conclusion.tex
\section{Conclusion}
\label{conclusion}

In this paper, we presented \netNameNew{}, a unified framework for motion forecasting and motion planning that decouples trajectory representations into motion modes and dynamic states. This formulation enables the model to explicitly capture both high-level directional intentions and fine-grained spatiotemporal motion progress. To effectively model these decoupled representations, we introduced three core modules that integrate Attention and Mamba mechanisms for robust scene understanding and temporally consistent prediction.
\ext{
We further enhanced the framework with cross-scene intention interaction and state anchor-based refinement, which significantly improve accuracy and robustness, particularly in complex and continuous driving scenarios.
Moreover, we extended the application of our framework beyond forecasting to planning tasks, including both conventional motion planning and end-to-end autonomous driving based on raw sensor inputs.
Extensive experiments on Argoverse 2, nuScenes, nuPlan, and NAVSIM benchmarks demonstrate that our approach achieves state-of-the-art performance consistently.
}

\textbf{Limitations and future work.}
The proposed framework adopts a decoupled query paradigm, which may lead to heavier models due to the need to predict longer trajectories. Our current model design does not sufficiently take model efficiency into account. In the future, we plan to use sparse states for modeling trajectories, thereby making the framework more deployment-friendly.